\documentclass{llncs}
\usepackage[dvips]{graphics}
\usepackage{epsfig}
\usepackage{amsfonts}
\usepackage{amsmath}
\usepackage{color}
\usepackage{url}
\usepackage{multicol}
\usepackage{times}
\usepackage{algorithmic,algorithm}

\begin{document}

\pagestyle{empty}

\mainmatter

\title{First-improvement vs. Best-improvement \\ Local Optima Networks of NK Landscapes}

\author{ Gabriela Ochoa\inst{1} \and S\'ebastien Verel\inst{2} \and Marco Tomassini\inst{3}}

\authorrunning{G. Ochoa, S. Verel, and M. Tomassini}   

\institute{School of Computer Science, University of Nottingham, Nottingham, UK. \and INRIA Lille - Nord Europe and University of Nice Sophia-Antipolis, France.
   \and Information Systems Department, University of Lausanne, Lausanne, Switzerland.}

\maketitle

\begin{abstract}
This paper extends a recently proposed model for combinatorial landscapes: {\em Local Optima Networks (LON)}, to incorporate a first-improvement (greedy-ascent) hill-climbing algorithm, instead of a best-improvement (steepest-ascent) one,   for the definition and extraction of the basins of attraction of the landscape optima. A statistical analysis comparing best and first improvement network models for  a set of $NK$ landscapes, is presented and discussed. Our results suggest structural differences between the two models with respect to both the network connectivity, and the nature of the basins of attraction. The impact of these differences in the behavior of search heuristics based on first and best improvement local search is thoroughly discussed.
\end{abstract}

\section{Introduction}
\label{sec:intro}

\begin{figure} [!ht]
\begin{center}

\includegraphics[width=0.5\textwidth]{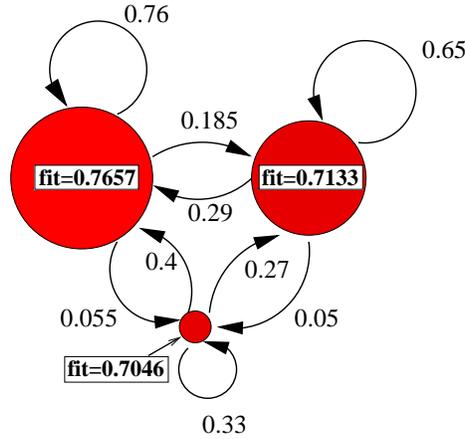} 
\vspace{-0.4cm} \caption{Visualisation of the weighted local optima network of a small $NK$ landscape ($N=6$, $K=2$). The nodes correspond to the local optima basins (with the diameter indicating the size of basins, and the label "fit", the fitness of the local optima). The  edges depict the transition probabilities between basins as defined in the text.  \label{fig:network}}
\end{center}
\end{figure}

The performance of heuristic search algorithms crucially depends on the structural aspects of the spaces being searched. An improved understanding of this dependency, can facilitate the design and further successful application of these methods to solve hard computational search problems. Local optima networks (LON) have been recently introduced as a novel model of combinatorial landscapes \cite{gecco08,pre08,tec10}. This model allows the use of  complex network analysis techniques \cite{newman03} in connection with the study of fitness landscapes and problem difficulty in combinatorial optimisation. The model, inspired by work in the physical sciences on energy surfaces \cite{doye02},  is based on the idea of compressing the information given by the whole problem configuration space into a smaller mathematical object which is the graph having as vertices the optima configurations of the problem and as edges the possible weighted transitions between these optima (see Figure \ref{fig:network}). This characterization of landscapes as networks has brought new insights into the global structure of the landscapes studied, particularly into the distribution of their local optima. Moreover, some network features have been found to correlate and suggest explanations for search difficulty on the studied domains. The study of local optima networks has also revealed new properties of the basins of attraction.

The current methodology for extracting LONs requires the exhaustive exploration of the search space, and the use of a best-improvement (steepest-ascent) local search algorithm from each configuration. In this paper, we are interested in exploring how the network structure and features of a given landscape will change, if a first-improvement  (greedy-ascent) local search algorithm is used instead for extracting the basins and transition probabilities.   This is apparently simple but, in reality, requires a careful redefinition of the concept of a basin of attraction. The new notions will be presented in the next section.  Following previous work \cite{pre08,tec10}, we use the well-known family of $NK$ landscapes \cite{kauffman93} as an example, as it allows the exploration of landscapes of tunable ruggedness and search difficulty.

The article is structured as follows. Section \ref{sec:defs}, includes the relevant definitions and algorithms for extracting the LONs. Section \ref{sec:analysis} describes the experimental design, and reports the analysis of the extracted networks, including a study of both their basic features and connectivity, and the nature of the basins of attraction of the local optima. Finally, section \ref{sec:discussion} discusses our main findings and suggest directions for future work.

\section{Definitions and algorithms}
\label{sec:defs}

A Fitness landscape~\cite{stadler-02} is a triplet $(S, V, f)$ where $S$ is a set of potential solutions i.e. a search space, $V : S \longrightarrow 2^S$, a neighborhood structure, is a function that assigns to every $s \in S$ a set of neighbors $V(s)$, and $f : S \longrightarrow R$ is a fitness function that can be pictured as the \textit{height} of the corresponding solutions. In our study, the search space is composed by binary strings of length $N$, therefore its size is $2^N$. The neighborhood is defined by the minimum possible move on a binary search space, that is, the 1-move or bit-flip operation. In consequence, for any given string $s$ of length $N$, the neighborhood size is $|V(s)| = N$. The $HillClimbing$ algorithm to determine the local optima and therefore define the basins of attraction, is given in Algorithm~\ref{algoHC}. It defines a mapping from the search space $S$ to the set of locally optimal solutions $S^*$.

First-improvement  differs from best-improvement local search, in the way of selecting the next neighbor in the search process, which is related with the so-called  {\em pivot-rule}. In best-improvement, the entire neighborhood is explored and the best solution is returned, whereas in first-improvement, a solution is selected uniformly at random  from the neighborhood (see  Algorithm ~\ref{algoHC}).

\begin{algorithm}[t]
\begin{multicols*}{2}
\begin{algorithmic}
\STATE Choose initial solution $s \in S$
\REPEAT
    \STATE choose $s^{'} \in V(s)$, such that $f(s^{'}) = max_{x \in V(s)} f(x)$
        \IF{$f(s) < f(s^{'})$}
            \STATE $s \leftarrow s^{'}$
    \ENDIF
\UNTIL{$s$ is a  Local optimum}
 \end{algorithmic}

\columnbreak

 \begin{algorithmic}
\STATE Choose initial solution $s \in S$
\REPEAT
    \STATE choose  $s^{'} \in V(s)$ using a predefined random ordering
    \STATE
        \IF{$f(s) < f(s^{'})$}
            \STATE $s \leftarrow s^{'}$
    \ENDIF
\UNTIL{$s$ is a  Local optimum}
 \end{algorithmic}

\end{multicols*}
\vspace{-0.3cm}
\caption{Best-improvement (left) and first-improvement (right) algorithms.}
\label{algoHC}
\end{algorithm}

First, let us define the standard notion of a local optimum.

\textbf{Local optimum (LO).} A local optimum, which is taken to be a maximum here, is a solution $s^{*}$ such that $\forall s \in V(s)$, $f(s) \leq f(s^{*})$.

Let us denote by $h$, the stochastic operator that  associates to each solution $s$, the solution obtained after applying  one of the hill-climbing algorithms (see Algorithms \ref{algoHC}) for a sufficiently large number of iterations to converge to a $LO$. The size of the landscape is finite, so we can denote by $LO_1$, $LO_2$, $LO_3 \ldots, LO_p$, the local optima. These $LOs$ are the vertices of the \textit{local optima network}.

Now, we introduce the concept of basin of attraction to define the edges and weights of our network model. Note that for each solution $s$, there is a probability that $h(s) = LO_{i}$. We denote $p_i(s)$ the probability $P(h(s) =  LO_{i})$. We have that for:
\begin{description}
\item [Best-improvement:] for a given  solution $s$, there is a (single) local optimum, and thus an $i$, such that $p_i(s)=1$ and $\forall j \neq i, p_j(s) = 0$.
\item [First-improvement:]  for a given solution $s$, it  is possible to have several local optima, and thus several $i_1, i_2, \ldots, i_ m$, such that $p_{i_1}(s) > 0, p_{i_2}(s) > 0, \ldots, p_{i_m}(s) > 0$.
\end{description}

For both models, we have, for each solution $s \in S$, $\sum_{i=1}^{n} p_i(s) = 1$.

Following the definition of the LON model in neutral fitness landscapes \cite{tec10}, we have that:

\textbf{Basin of attraction.} The basin of attraction of the local optimum $i$ is the set $b_i = \{ s \in S ~|~ p_i(s) > 0 \}$. This definition is consistent with our previous definition \cite{pre08} for the best-improvement case.\\

The size of the  basins of attraction can now be defined as follows:

\textbf{Size of a basin of attraction.} The size of the basin of attraction of a local optimum  $i$ is $\sum_{s \in {\cal S}} p_i(s)$.\\

\textbf{Edge weight.} We first reproduce the definition of edge weights for the non-neutral landscape, and best-improvement hill-climbing \cite{pre08}:
\noindent For each solutions $s$ and $s^{'}$, let  $p(s \rightarrow s^{'} )$ denote the probability that $s^{'}$ is a neighbor
of $s$, \textit{i.e.} $s^{'} \in V(s)$. Therefore, we define below: $p(s \rightarrow b_j )$,  the probability that a configuration $s \in S$ has a neighbor in a basin $b_j$, and $p(b_i \rightarrow b_j)$,   the total probability of going from basin $b_i$ to basin $b_j$, which is as the average over all $s \in b_i$ of the transition probabilities to solutions $s^{'} \in b_j$ (where $\sharp b_i$  is the size of the basin $b_i$) :
$$ p(s \rightarrow b_j ) = \sum_{s^{'} \in b_j} p(s \rightarrow s^{'} ), \hspace{1.2cm} p(b_i \rightarrow b_j) = \frac{1}{\sharp b_i} \sum_{s \in b_i}  p(s \rightarrow b_j )$$

For first and best improvement hill-climbing, we have defined the probability $p_i(s)$ that a solution $s$ belongs to a basin $i$.  We can, therefore, modify the previous definitions to consider both types of network models:

$$ p(s \rightarrow b_j ) = \sum_{s^{'} \in b_j} p(s \rightarrow s^{'} ) p_{j}(s^{'}),  \hspace{1.2cm}  p(b_i \rightarrow b_j) = \frac{1}{\sharp b_i} \sum_{s \in b_i} p_i(s) p(s \rightarrow b_j )$$

\noindent In the best-improvement, we have $p_k(s)=1$ for all the configurations in the basin $b_k$. Therefore, the definition of weights for the best-improvement case is consistent with the previous definition. Now, we are in a position to define the weighted local optima network:

\textbf{Local optima network.} The weighted local optima network $G_w=(N,E)$ is the graph where the nodes are the local optima, and there is an edge $e_{ij} \in E$, with  weight $w_{ij} = p(b_i \rightarrow b_j)$, between two nodes $i$ and $j$ if $p(b_i \rightarrow b_j) > 0$.

According to our definition of edge weights, $w_{ij} = p(b_i \rightarrow b_j)$ may be different than $w_{ji} = p(b_j \rightarrow b_i)$. Thus, two weights are needed in general, and we have an oriented transition graph.

\section{Analysis of the local optima networks}
\label{sec:analysis}

The $NK$ family of landscapes \cite{kauffman93} is a problem-independent model for constructing multimodal landscapes that can gradually be tuned from smooth to rugged. In the model, $N$ refers to the number of (binary) genes in the genotype (i.e. the string length) and $K$ to the number of genes that influence a particular gene. By increasing the value of $K$ from 0 to $N-1$, $NK$ landscapes can be tuned from smooth to rugged. The $K$ variables that form the context of the fitness contribution of gene $s_i$ can be chosen according to different models. The two most widely studied models are the {\em random neighborhood} model, where the $K$  variables are chosen randomly according to a uniform distribution among the $n-1$ variables other than $s_i$, and the {\em adjacent neighborhood} model, in which the $K$ variables that are closest to $s_i$ in a total ordering $s_1, s_2, \ldots, s_n$ (using periodic boundaries). No significant differences between the two models were found in \cite{kauffman93} in terms of  the landscape global properties, such as mean number of local optima or autocorrelation length. Similarly, our preliminary studies on the characteristics of the $NK$ landscape optima networks, did not show noticeable differences between the two neighborhood models. Therefore, we conducted our full study on the more general random model.

In order to minimize the influence of the random creation of landscapes, we considered 30 different and independent landscapes for each combination of  $N$ and $K$ parameter values. In all cases, the measures reported, are the average of these 30 landscapes. The study considered landscapes with  $N \in \{ 14, 16 \}$ and $K \in \{2,4, \ldots, N-1\}$, which are the largest possible parameter combinations  that allow the exhaustive extraction of local optima networks. Both best-improvement and first-improvement local optima networks (b-LON and f-LON, respectively) were extracted and analyzed.

\subsection{Network features and connectivity}
This section reports the most commonly used features to characterise complex networks, in both the f-LON and b-LON models.

\begin{table}[!ht]
\begin{center}
\small \caption{$NK$ landscapes network properties.  Values are
averages over 30 random instances, standard deviations are shown as
subscripts. $n_v$ and $n_e$ represent the number of vertexes and
edges, $\bar C^{w}$, the mean weighted
clustering coefficient. $\bar Y$ represent the mean disparity
coefficient, $\bar d$ the mean path length, and $\bar d_{best}$ the mean path length to the global optimum (see text for
definitions). } \label{tab:statistics} \vspace{0.2cm}

\begin{tabular}{|c|c|c|c|c|c|c|c|c|c|c|}
\hline
$K$ & $\bar n_v$ & \multicolumn{2}{|c|}{$\bar n_e / {\bar n_v}^2$} & $\bar C^{w}$ & \multicolumn{2}{|c|}{$\bar Y$} & \multicolumn{2}{|c|}{$\bar d$} & \multicolumn{2}{|c|}{$\bar d_{best}$} \\
\hline
\multicolumn{11}{|c|}{$N = 14$} \\
\hline
   & both & b-LON & f-LON & b-LON & b-LON & f-LON & b-LON & f-LON & b-LON & f-LON \\
\hline
2  &    $14_{ 6}$  & $0.89$ & $1.00$  & $0.98_{0.015}$  &  $0.367_{0.0934}$  & $0.172_{0.0977}$ & $76_{194}$  & $28_{18}$   & $13_{6}$    & $10_{6}$  \\
4  &    $70_{10}$  & $0.64$ & $1.00$  & $0.92_{0.013}$  &  $0.148_{0.0101}$  & $0.048_{0.0079}$ & $89_{6}$    & $86_{7}$    & $26_{8}$    & $23_{11}$ \\
6  &   $184_{15}$  & $0.37$ & $1.00$  & $0.79_{0.014}$  &  $0.093_{0.0031}$  & $0.025_{0.0017}$ & $119_{3}$   & $140_{6}$   & $44_{9}$    & $49_{16}$ \\
8  &   $350_{22}$  & $0.21$ & $1.00$  & $0.66_{0.015}$  &  $0.070_{0.0020}$  & $0.017_{0.0008}$ & $133_{2}$   & $183_{4}$   & $67_{10}$   & $95_{20}$ \\
10 &   $585_{22}$  & $0.12$ & $1.00$  & $0.54_{0.009}$  &  $0.058_{0.0010}$  & $0.014_{0.0004}$ & $139_{1}$   & $218_{3}$   & $84_{11}$   & $141_{26}$\\
12 &   $896_{22}$  & $0.07$ & $1.00$  & $0.46_{0.004}$  &  $0.052_{0.0006}$  & $0.013_{0.0002}$ & $140_{1}$   & $247_{2}$   & $102_{11}$  & $196_{42}$\\
13 &  $1,085_{20}$ & $0.06$ & $1.00$  & $0.42_{0.004}$  &  $0.050_{0.0006}$  & $0.013_{0.0002}$ & $139_{1}$   & $259_{1}$   & $104_{9}$   & $218_{38}$\\

\hline \hline
\multicolumn{11}{|c|}{$N = 16$} \\
\hline
   & both & b-LON & f-LON & b-LON & b-LON & f-LON & b-LON & f-LON & b-LON & f-LON \\
\hline
2  &     $33_{15}$ &  $0.81$  &  $1.00$  & $0.96_{0.024}$  & $0.326_{0.0579}$  & $0.110_{0.0590}$ & $56_{14}$  &  $39_{11}$   & $16_{5}$    & $12_{5}$  \\
4  &    $178_{33}$ &  $0.60$  &  $1.00$  & $0.92_{0.017}$  & $0.137_{0.0111}$ & $0.033_{0.0064}$ & $126_{8}$  &  $127_{13}$   & $35_{9}$    & $32_{13}$ \\
6  &    $460_{29}$ &  $0.32$  &  $1.00$  & $0.79_{0.015}$  & $0.084_{0.0028}$ & $0.016_{0.0014}$ & $170_{3}$  &  $215_{8}$    & $60_{15}$   & $70_{23}$ \\
8  &    $890_{33}$ &  $0.17$  &  $1.00$  & $0.65_{0.010}$  & $0.062_{0.0011}$ & $0.011_{0.0004}$ & $194_{2}$  &  $282_{5}$    & $83_{13}$   & $118_{26}$\\
10 &  $1,470_{34}$ &  $0.09$  &  $1.00$  & $0.53_{0.007}$  & $0.050_{0.0006}$ & $0.009_{0.0002}$ & $206_{1}$  &  $340_{3}$    & $112_{15}$  & $183_{30}$\\
12 &  $2,254_{32}$ &  $0.05$  &  $1.00$  & $0.44_{0.003}$  & $0.043_{0.0003}$ & $0.008_{0.0001}$ & $207_{1}$  &  $380_{2}$    & $143_{16}$  & $271_{48}$\\
14 &  $3,264_{29}$ &  $0.03$  &  $1.00$  & $0.38_{0.002}$  & $0.040_{0.0003}$ & $0.008_{0.0001}$ & $203_{1}$  &  $411_{1}$    & $158_{13}$  & $351_{51}$\\
15 &  $3,868_{33}$ &  $0.02$  &  $1.00$  & $0.35_{0.002}$  & $0.039_{0.0004}$ & $0.008_{0.0000}$ & $200_{1}$  &  $423_{1}$    & $162_{13}$  & $391_{87}$\\

\hline
\end{tabular}
\end{center}
\end{table}

\begin{figure}[!ht]
\begin{center}
\begin{tabular}{cc}
 $W_{ii}$  & $W_{ij}$\\
\includegraphics[width=0.5\textwidth]{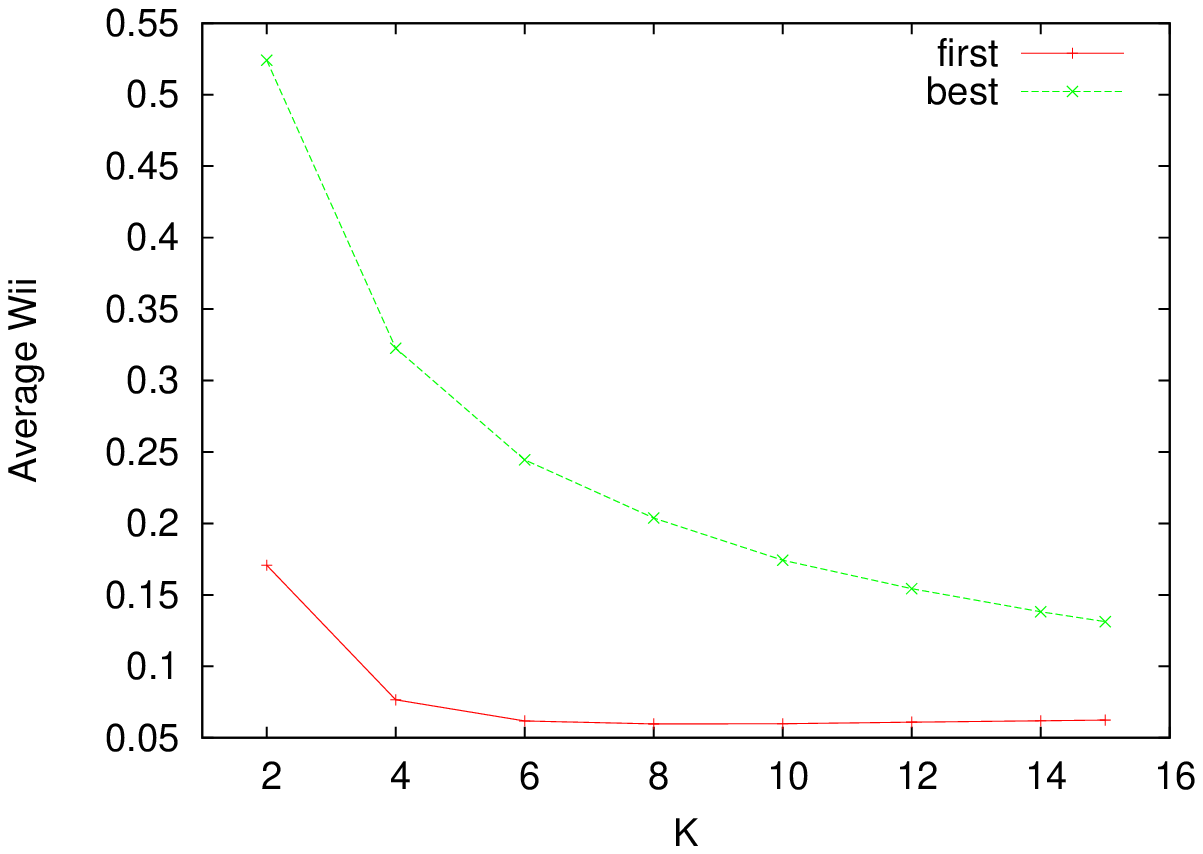} &
\includegraphics[width=0.5\textwidth]{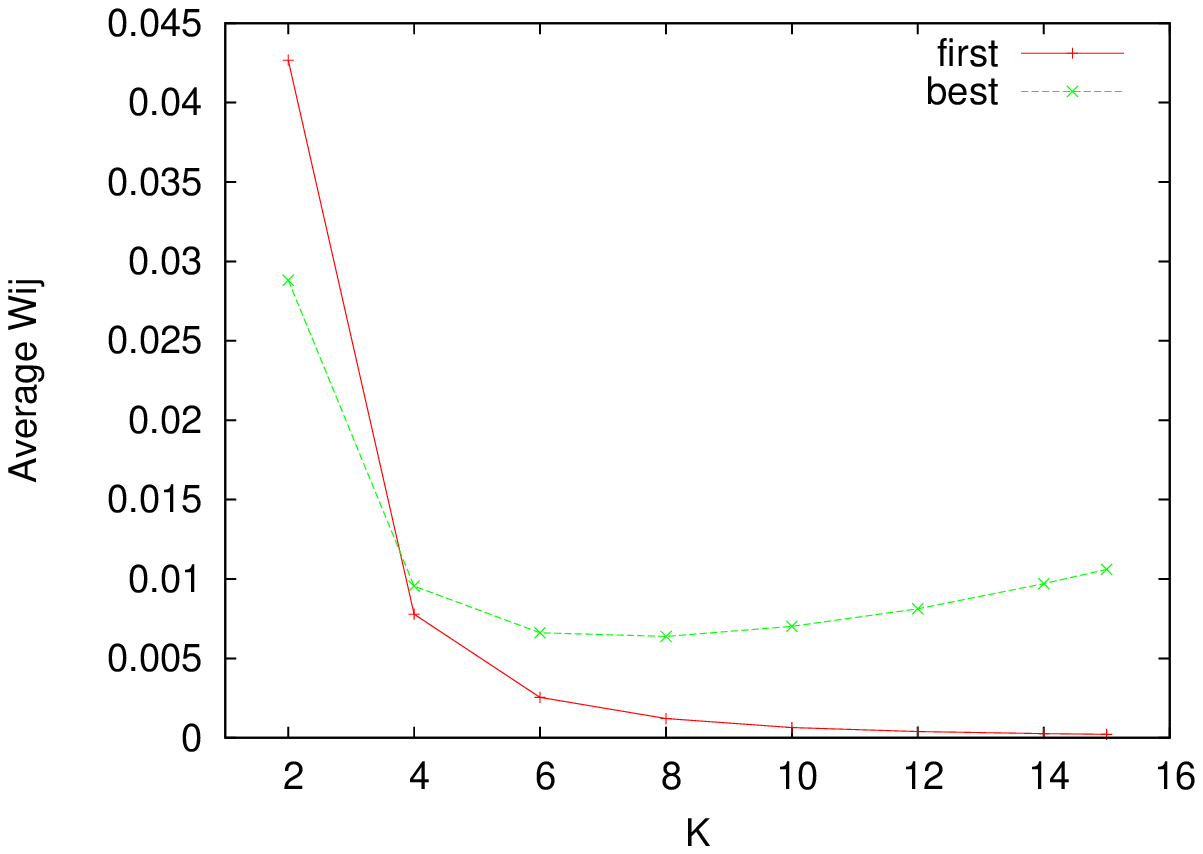} \\
\end{tabular}
\vspace{-0.3cm}
\caption{  Averages of  $w_{ii}$ weights (left), and   averages of   $w_{ij}$ with $j \not= i$ weights (right), for landscapes with $N=16$ and all the $K$ values. \label{fig:weigths}}
\end{center}
\end{figure}

{\bf  Number of nodes and edges}:  The $2^{nd}$ column of Table \ref{tab:statistics}, reports the number of nodes (local optima),$n_v$, for all the studied landscapes. The b-LONs and f-LONs  have  the same local optima, since both local search algorithms, although using a different pivot-rule,  are  based on the bit-flip neighborhood. The networks, however, have a different number of edges, as can be appreciated in the $3^{rd}$ and $4^{th}$ columns of  Table \ref{tab:statistics}, which report  the number of edges normalized by the square of the number of nodes. Clearly, the number of edges is much larger for the f-LONs. This number is always the square of the number of nodes, which indicates that  the f-LONs are  complete graphs. It is worth noticing, however, that many of the edges have very low weights (see Figure \ref{fig:weigths}). For the b-LON model,  the  number of edges decrease steadily with increasing values of $K$.

{\bf Clustering coefficient or transitivity}: The {\em clustering coefficient} of a network is the average probability that  that two neighbors of a given node are also neighbors of  each other. In the language of social networks,  the friend of your friend is likely also to be your friend. The standard  clustering coefficient \cite{newman03} does not consider weighted edges. We thus used the {\em weighted clustering} measure proposed by \cite{bart05}. The $5^{th}$ column of table \ref{tab:statistics} lists the average coefficients of the b-LONs for all $N$ and $K$. It is apparent that the clustering coefficients decrease regularly with increasing $K$, which indicates that either there are less transitions between neighboring basins for high $K$, and/or the transitions are less likely to occur.  On the other hand,  the f-LONs correspond to complete networks; the calculation of the clustering coefficients revealed  that  $\forall i$, $c^{w}(i) = 1.0$ (not shown in the Table). Therefore, the f-LON is densely connected for all values of $K$.

{\bf Disparity}:  The {\em disparity} measure proposed in \cite{bart05},  $Y(i)$,  gauges the heterogeneity of the contributions of the edges of node $i$ to the total weight. Columns $6^{th}$ and $7^{th}$ in Table \ref{tab:statistics} depict the disparity coefficients, for both  network models, respectively. The heterogeneity decreases with increasing values of $K$. This reflects that with high values of $K$, the transitions to other basins tend to become equally likely, an indication of a more random structure (and thus a difficult search). It can also be seen that the weights for the f-LON model are less heterogenous (more uniform) than for the b-LON one.

{\bf Shortest path length}: Another standard metric to characterize the structure of networks is the shortest path length (number of link hobs) between two nodes on the network. In order to compute this measure on the optima network of a given landscape, we considered the expected number of bit-flip mutations to pass from one basin to the other. This expected number can be computed by considering the inverse of the transition probabilities between basins.  More formally, the distance between two nodes is defined by $d_{ij} = 1 / w_{ij}$ where $w_{ij} = p(b_i \rightarrow b_j)$. Now, we can define the length of a path between two nodes as being the sum of these distances along the edges that connect the respective basins. Columns $9^{th}$ and  $7^{th}$ in Table \ref{tab:statistics} report this measure on the two network models.  In both cases, the shortest path increases with $K$, however, for the b-LON the growth stagnates for larger $K$ values. The paths are considerably longer for the f-LON, with the exception of the lowest values of $K$. Some paths are more relevant from the point of view of a stochastic local search algorithm following a trajectory over the maxima network. Therefore, columns $10^{th}$ and  $11^{th}$ in Table \ref{tab:statistics}, report the shortest path length to the global optimum from all the other optima in the landscape. The trend is clear, the path lengths to the optimum increase steadily with increasing $K$, and similarly, the first-improvement network shows longer paths. This suggest that a larger number of hops  will be needed to find the global optimum when a first-improvement local search is used. We must consider, however, that the number of evaluations needed to explore a basin, would be $N$ times lower for first-improvement than for best-improvement.

{\bf Outgoing weight distribution}: The standard topological characterization of (unweighed) networks is obtained by its degree distribution. The degree of a  node is defined as its number of neighbours, and the  degree distribution of a network is the distribution over the frequencies of different degrees over all nodes in the network. For  weighted networks, a characterization of weights is obtained by the {\em connectivity and weight distributions} $p_{in}(w)$ and $p_{out}(w)$ that any given edge has incoming or outgoing weight $w$. In our study, for each node $i$, the sum of outgoing edge weights is equal to $1$ as they represent transition probabilities. So, an important measure is the weight $w_{ii}$ of self-connecting edges (remaining in the same node). We have the relation: $ w_{ii} + s_i = 1$.

\begin{figure}[!ht]
\centering
\includegraphics[width=0.6\textwidth]{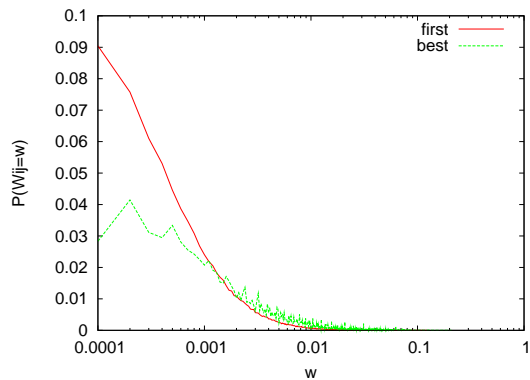} \\
\vspace{-0.3cm}
\caption{Probability distribution of the network weights $w_{ij}$ for outgoing edges with $j\not=i$ (in logscale on x-axis) for  $N=16, K=6$. Averages on 30 independent landscapes.\label{fig:Wij_distri}}
\end{figure}

Figure \ref{fig:Wij_distri}, reports the outgoing weight distributions $p_{out}(w)$ (in log-scale on x-axis) of both the f-LON and b-LON networks on a selected landscape with  $K=6$, and $N=16$.  One can see that the weights, i.e. the transition probabilities to neighboring basins are small. The distributions are far from uniform or Poissonian, they are not close to power-laws either. We couldn't find a simple fit to the curves such as stretched exponentials or exponentially truncated power laws.  It can be seen that the distributions differ for the first and best LON models. There is  a larger number of edges with low weights for the f-LONs than for the b-LONs. Thus, even though the f-LONs are more densely connected (indeed they are complete graphs) many of the edges have very low weights. Figure \ref{fig:weigths} (left), shows the averages, over all the nodes in the network, of the weights $w_{ii}$ (i.e. the probabilities of remaining in the same basin after a bit-flip mutation) for $N=16$ and all the $K$ values. Notice that, for both network models,  the weights $w_{ii}$ are much higher when  compared to those $w_{ij}$ with $j \not= i$ (see Fig.~\ref{fig:weigths} right). The $w_{ii}$ are much lower for the first than for the best LON. In particular, in the  b-LON, for $K=2$, $50\%$ of the random bit-flip mutations will produce a solution within the same basin of attraction, whereas this figure is of less than $20\%$ in the f-LON. Indeed, in this case, for $K$ greater than 4, the probabilities of remaining in the same basin fall below  $10\%$, which suggests that escaping  from local optima would be easier for a first-improvement local searcher.

\subsection{Basins of attraction features}
The previous section studied and compared the basic network features and connectivity of the first and best LONs. The exhaustive extraction of the networks, also produced detailed information of the corresponding basins of attraction.  Therefore, this section discusses the most relevant  of the basin's features.

{\bf Size of the global optimum basin}: When exploring the average size of the global optimum basin of the f-LONs, we found that they decrease exponentially with increasing ruggedness ($K$ values). This is consistent with the results for the b-LON on these landscapes \cite{pre08}. Moreover, the basins sizes for both networks are similar, with those of f-LON being slightly smaller. This may suggest that for the the same number of runs, the success rate of a first-improvement heuristic would be lower. One needs to consider, however, that the number of evaluations per run is smaller in this case.

{\bf Basin sizes of the two network models}: A comparative study of the basin sizes of the two network models revealed that they are highly correlated. Only the smallest basins of the f-LON model are larger in size when compared to the corresponding smallest basins in the b-LON model.

\begin{figure} [!ht]
\begin{center}
\includegraphics[width=0.6\textwidth]{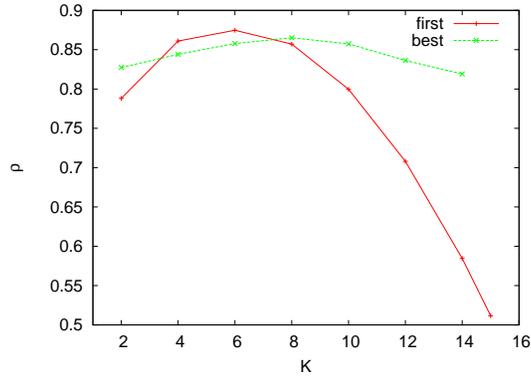} 
\end{center}
\vspace{-0.3cm}
\caption{Average of the correlation coefficient between the fitness of local optima and their corresponding basin sizes on $30$ independent landscapes for both f-LON and b-LON  ($N=16$, and all the $K$ values).\label{fig:cor_fit-size}}
\end{figure}

{\bf Basin size and fitness of local optima}: Fig.~\ref{fig:cor_fit-size} reports the correlation coefficients $\rho$ between the networks' basin sizes and their fitness, for both the first and best LONs, and landscapes with $N=16$ and all the $K$ values. It can be observed that there is a strong correlation between fitness and basin sizes for both types of networks.  Indeed, for $K \leq 10$, the correlation is over $\rho > 0.8$. For rugged landscapes, $K > 8$, the  f-LON shows reduced and decreasing coefficients as compared to  the b-LON.

\begin{figure*} [!ht]
\begin{center}
\begin{tabular}{cc}
(a) & (b) \\
\includegraphics[width=0.5\textwidth]{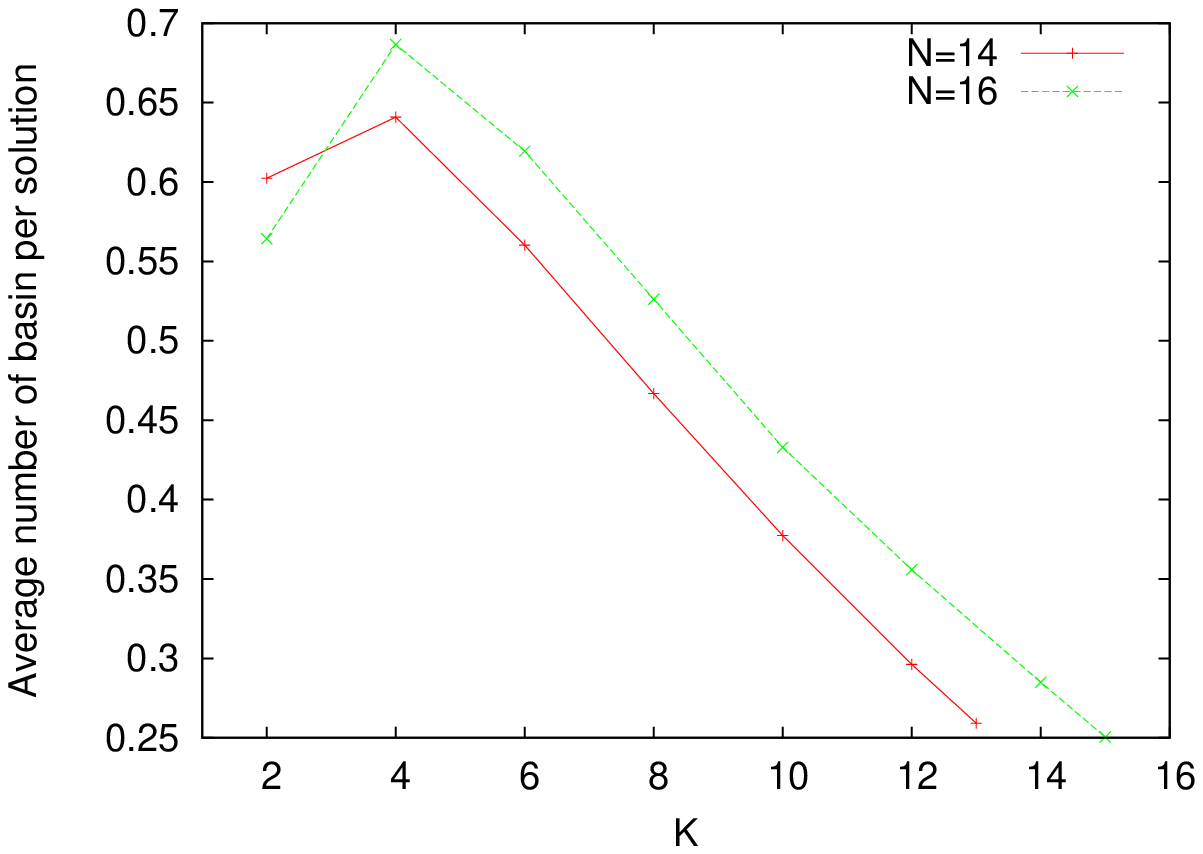} &
\includegraphics[width=0.5\textwidth]{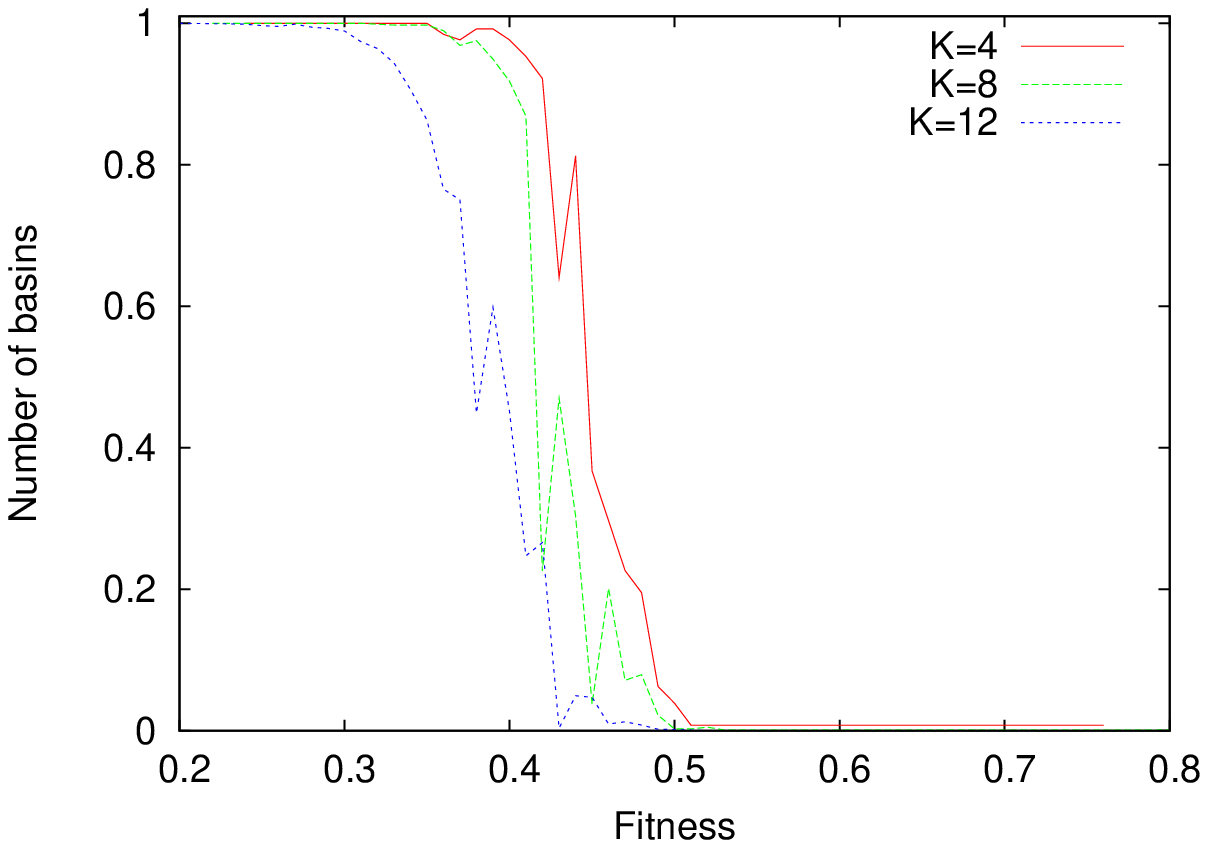} \\
\end{tabular}
\end{center}
\vspace{-0.8cm}
\caption{(a) Average number of basins to which a solution belongs. (b) For $N=16$ and 3 selected values of $K$, the number of basins per solution according to the solution fitness value. Averages on 30 independent landscapes.\label{fig:nbBasinSol}}
\end{figure*}

{\bf Number of basins per solution on the f-LONs}: According to the definition of basins  (see section \ref{sec:defs}), for  the f-LON,  a given solution may belong to a set of basins. Fig. \ref{fig:nbBasinSol} (a) shows the average number of basins to  which a solution belongs  (i.e.  $\sharp \{ i ~|~ p_i(s) > 0 \}$). It can be observed that for  $N=16$ and $K=4$, a solution belongs to nearly $70\%$ of the total number of basins, whereas for $K=14$, a solution belongs to less than $30\%$ of the total number of basins. On average, a solution belongs to less basins for high $K$ than for low $K$.  An exploration of the average number of  basin per solution, according to the solution fitness value (Fig. \ref{fig:nbBasinSol} (b), for $N=16$) reveals a striking difference. While low fitness solutions belong to nearly all basins, high fitness solutions belong to at most one basin. The figure suggest the presence of a phase transition, in which the threshold of the transition is lower for high $K$ than for low $K$. This suggests that the structure of the f-LON network for solutions with high fitness, resembles that of the b-LON, whereas the topology is different with respect to solutions with low fitness.

\vspace{-0.3cm}

\section{Discussion}
\label{sec:discussion}
We have extended the recently proposed {\em Local Optima Network  (LON)} model to analyze the structural differences between first and best improvement local search, in terms of the local optima network connectivity and the nature of the corresponding basins of attraction. The results of the analysis, on a set of $NK$ landscapes can be summarized as follows. The impact of landscape ruggedness ($K$ value) on the network features is similar for both models. First-improvement induces a densely connected network (indeed a complete network), while this is not the case on the best-improvement model. However, many of the edges in the f-LON networks have very low weights. In particular, the self-connections (i.e. the probabilities of remaining in the same basin after a bit-flip mutation), are much smaller in the f-LON than in the b-LON model, which suggests  that escaping  from local optima would be easier for a first-improvement local searcher. The path lengths between local optima, and between any optima and the global optimum, are generally larger in f-LON than in b-LON networks. We must consider, however, that the number of evaluations needed to explore a basin, would be $N$ times lower for first-improvement than for best-improvement. We, therefore, suggest that first-improvement is a better heuristic for exploring $NK$ landscapes. Our preliminary empirical results support this insight, a  detailed account of them will be presented elsewhere due to space restrictions.
Most of our work on the local optima model has been based on binary spaces and $NK$ landscapes. However, we have recently started the exploration of permutation search spaces, specifically the Quadratic Assignment Problem (QAP) \cite{cec10}, which opens up the possibility of analyzing other permutation based problems such as the traveling salesman and the permutation flow shop problems. Our current definition of transition probabilities, although very informative,  produces highly connected networks, which are not easy to study. Therefore, we are currently considering alternative definitions and threshold values for the connectivity. Finally, although the local optima network model is still under development, we argue that it offers an alternative view of combinatorial fitness landscapes, which can potentially contribute to both  our understanding of problem difficulty, and the design of effective heuristic search algorithms.
\bibliographystyle{plain}
\small

\end{document}